\RequirePackage{amsmath}
\documentclass[]{llncs}
\pdfoutput=1

\usepackage[utf8]{inputenc} 
\usepackage[T1]{fontenc}    

\usepackage{amssymb}
\usepackage{graphicx}

\usepackage{hyperref}
\usepackage{mathtools}
\usepackage{booktabs}
\usepackage{siunitx}
\usepackage{xcolor}

\DeclarePairedDelimiter\abs{\lvert}{\rvert}
\DeclarePairedDelimiter\norm{\lVert}{\rVert}

\begin{document}

\title{Singular Value Decomposition and Neural Networks}

\author{%
	Bernhard Bermeitinger\inst{1,2}\orcidID{0000-0002-2524-1850}
	\and Tomas Hrycej\inst{1}
	\and \\ Siegfried Handschuh\inst{1,2}}


\institute{Chair of Data Science,
	Institute of Computer Science\\
	University of St.Gallen, St.Gallen, Switzerland\\
	\email{\{bernhard.bermeitinger,tomas.hrycej,siegfried.handschuh\}@unisg.ch}
	\and
	University Passau, Passau, Germany\\
	\email{\{bernhard.bermeitinger,siegfried.handschuh\}@uni-passau.de}
}

\maketitle

\setcounter{footnote}{0}

\begin{abstract}
Singular Value Decomposition (SVD) constitutes a bridge between the linear algebra concepts and multi-layer neural networks---it is their linear analogy. Besides of this insight, it can be used as a good initial guess for the network parameters, leading to substantially better optimization results.
\keywords{Singular Value Decomposition \and Neural Network \and Deep Neural Network \and Initialization \and Optimization \and Conjugate Gradient}
\end{abstract}

\section{Motivation}\label{sec:introduction}
The utility of multi-layer neural networks is frequently being explained by their capability of extracting meaningful features in their hidden layers. This view is particularly appropriate for large size applications such as corpus-based semantics analyses where the number of training examples is too low for making the problem fully determined in terms of a direct mapping from the input to the output space.

This capability of feature extraction is mostly implicitly attributed to using nonlinear units in contrast to a linear mapping. The prototype of such linear mapping is linear regression, using multiplication of an input pattern by a regression matrix to get an estimate of the output pattern, omitting the possibility of using a sequence of two (or more) matrices corresponding to the use of a hidden layer of linear units. This possibility is usually considered to be obsolete with the argument that a product of two matrices is also a matrix and the result is thus equivalent to using a single matrix.

This argument, although superficially correct, hides the possibility of using a matrix of deliberately chosen low rank, which leads to the correct treatment of under-determined problems.

A key to understanding the situation is Singular Value Decomposition (SVD)\@. In the following, it will be shown that SVD can be interpreted as a linear analogy of a neural network with one hidden layer and that it can be used for generating a good initial solution for optimizing nonlinear multi-layer neural networks. 

\section{Singular Value Decomposition}\label{sec:svd}
SVD is a powerful concept of linear algebra. It is a decomposition of an arbitrary matrix $ A $ of size $ m \times n $ into three factors:

\begin{equation}\label{Eq.SVD}
    A = U S V^T
\end{equation}

where $ U $ and $ V $ are orthonormal and $ S $ is of identical size as $ A $, consisting of a diagonal matrix $ D_0 $ and a zero matrix. For $ m < n $, it is $ \left[ S_0, 0 \right] $, for $ m > n $ it is $ \left[ S_0, 0 \right]^T $. In the further discussion, only the case of $ m < n $ will be considered as the opposite case is analogous.

SVD is then simplified to

\begin{equation}\label{Eq.SVDecon}
    A = U S V^T = U \left[ S_0, 0 \right] \left[ V_0, V_x \right]^T = U S_0 V_0^T
\end{equation}
by omitting redundant zero terms. This form is sometimes called \emph{economical}.

For the economical form (\ref{Eq.SVDecon}), the decomposition with $ r = \min{ \left( m, n \right) } $ has $ m r + r + n r = \left( m + n + 1 \right) r $ nonzero parameters. The orthonormality of $ U $ and $ V $ imposes $ 2 r $ unity norm constraints, and $ r \left( r - 1 \right) $ orthogonality constraints, resulting in a total number of 
\begin{equation}
    2 r + r \left( r - 1 \right) = r^2 + r
\end{equation}
constraints.

The number of free parameters amounts to 
\begin{equation}
    \left( m + n + 1 \right) r - r^2 - r
\end{equation}

which is

\begin{equation}
    \left( m + n + 1 \right) m - m^2 - m = m n, \quad \text{for } m < n
\end{equation}

and, analogically,

\begin{equation}
    \left( m + n + 1 \right) n - n^2 - n = m n, \quad \text{for } m > n
\end{equation}

So, the economical form of SVD possesses the same number of free parameters as the original matrix $ A $.

The number of nonzero singular values in $ S_0 $ is equal to the rank $ r $ of matrix $ A $. An interesting case arises if the matrix $ A $ is not full rank, that is, if ${ r < \min{\left( m, n \right)}}$. Then, some diagonal elements of $ S_0 $ are zero.
Reordering the diagonal elements of $ S_0 $ (and, correspondingly the columns of $ U $ and $ V_0 $) so that its nonzero elements are in the field $ S_1 $ and zero elements in $ S_2 $, the decomposition further collapses to 
\begin{equation}\label{Eq.SVDmin}
A = U S_0 V_0^T = \left[ U_1, U_2 \right]
    \left[
    {
        \begin{array}{cc}
            S_1 &   0 \\
              0 & S_2 \\
        \end{array}
   }
   \right]
\left[ V_1, V_2 \right]^T
 = U_1 S_1 V_1^T
\end{equation}
Then, with the help of orthogonality of $ U_1 $ and $ V_1 $, the matrix can be decomposed into the sum
\begin{equation}\label{Eq.SVDsum}
    A = U S V^T = \sum_{k=1}^{r}{s_k u_k v_k^T}
\end{equation}

An important property of SVD is its capability for a matrix approximation by a matrix of lower rank. In analogy to the partitioning the singular values with the help of $ S_1 $ and $ S_2 $ to nonzero and zero ones, they can be partitioned to large and small ones. Selecting the $ \hat{r} $ largest singular values makes (\ref{Eq.SVDsum}) to an approximation $ \hat{A} $ of matrix $ A $. This approximation has the outstanding property of being that with the minimum $ L_2 $ matrix norm of the difference $ \hat{A} - A $ 
\begin{equation}\label{Eq.BdiffL2}
    \norm*{\hat{A} - A}_2
\end{equation}
out of all matrices of rank $ \hat{r} $. 

The $ L_2 $ matrix norm of $ M $ is defined as an induced norm by the $ L_2 $ vector norm, so that it is defined as
\begin{equation}\label{Eq.L2def}
    \norm[\big]{M}_2 = \max_x{\frac{\norm[\big]{M x}_2}{\norm[\big]{x}_2}}
\end{equation}

In many practical cases, a relatively small number $ \hat{r} $ leads to approximations very close to the original matrix.
Equation~\ref{Eq.SVDsum} shows that this property can be used for an economical representation of a $ m \times n $ matrix $ A $ by only $ \hat{r} \left( m + n + 1 \right)$ numerical values.
The optimum approximation property is shown below to be relevant for the mapping approximation discussed below.

A further important application of SVD is an explicit formula for a matrix pseudo-inverse. Pseudo-inverse $ A^{+} $ is the analogy of an inverse matrix for the case of non-square matrices, with the property
\begin{equation}\label{Eq.PseudoInvProperty}
    A A^{+} A = A
\end{equation}

It can be easily computed with the help of SVD\@:
\begin{equation}\label{Eq.PseudoInv}
    X^{+} = V S_{\text{inv}}^T U^T
\end{equation}

with $ S_{\text{inv}} $ being a matrix of the same dimension as $ S $ with inverted non-zero elements $ \frac{1}{s_{ii}} $ on the diagonal.

\section{SVD and Linear Regression}\label{sec:svd_and_lr}

One of the applications of the pseudo-inverse (\ref{Eq.PseudoInvProperty}) is a computing scheme for least squares. The linear regression problem is specified by input/output column vector pairs $ \left( x_i, y_i \right) $, seeking the best possible estimates
\begin{equation}
    \hat{y}_i = B x_i + a
\end{equation}

in the sense of least squares.

The bias vector $ a $ can be received by extending the input patterns $ x_i $ by a unity constant. For simplicity, it will be omitted in the ongoing discussion.

The solution amounts to solving the equation 
\begin{equation}
    Y = B X
\end{equation}
with matrices $ Y $ and $ X $ made of the corresponding column vectors.
The optimum is found with the help of the pseudo-inverse $ X^{+} $ of $ X $.
In the over-determined case (typical for linear regression), the least squares solution is
\begin{equation}\label{Eq.RegrOD}
    B = Y X^{+} = Y X^T \left( X X^T \right)^{-1}
\end{equation}

In the under-determined case, there is an infinite number of solutions with zero approximation error. The following solution has the minimum matrix norm of $ B $:
\begin{equation}\label{Eq.RegrUD}
    B = Y X^{+} = Y \left( X^T X \right)^{-1} X^T
\end{equation}
Both (\ref{Eq.RegrOD}) and (\ref{Eq.RegrUD}) use the pseudo-inverse that can be easily computed with help of SVD according to (\ref{Eq.PseudoInv}).

\section{SVD and Mappings of a Given Rank}
Both the full SVD (\ref{Eq.SVD}) and its reduced rank form (\ref{Eq.SVDmin}) are products of a dense matrix $ U $, a partly or fully diagonal matrix $ S $, and a dense matrix $ V^T $. This suggests the possibility of viewing them as a product of two dense matrices $ U S $ and $ V^T $, or $ U $ and $ S V^T $. All these matrices are full rank, even if the original matrix $ B $ was not due to the under-determination.

The product $ U S $ and $ V^T $ is the sequence of two linear mappings. The latter matrix maps the $ n $-dimensional input space to an intermediary space of dimension $ \hat{r} $, the former the intermediary space to the $ m $-dimensional output space. Since $ n > \hat{r} $ and $ m > \hat{r} $, the intermediary space represents a bottleneck similar to a hidden layer of a neural network. The orthogonal columns of $ V $ can be viewed as hidden features compressing the information in the input space.
This relationship to neural networks be followed in Section~\ref{sec:svd_and_lnn}.

The reasons to search for such a compressed mapping are different for the over-determined and the under-determined problems.

\subsection{Over-determined Problems}\label{sec:overdetermined_problems}
Suppose for an over-determined problem with input matrix $ X $ and output matrix $ Y $, the best linear solution is sought. The columns of $ X $ and $ Y $ correspond to the training examples. The least-square-optimum solution is the linear regression
\begin{equation}
    y = B x + a
\end{equation}
with matrix $ B $ from (\ref{Eq.RegrOD}).
The bias vector $ a $ can be received by extending the matrix $ X $ by a unit row and applying the pseudo-inversion of such an extended matrix. The last column of such an extended regression matrix corresponds to the column bias vector $ a $.

The linear regression matrix is $ m \times n $ for input dimension $ n $ and output dimension $ m $, its SVD is as in (\ref{Eq.SVD}).

With more than $ n $ independent training examples, the regression matrix $ B $ and also the matrix $ S $ are full rank with singular values on the diagonal of $ S $. 

There may be reasons for assuming that there are random data errors, without which the rank of $ B $ would not be full. This would amount to the assumption that some of the training examples are, in fact, linearly dependent or even identical and only the random data errors make them different. To ensure correct generalization, it would then be appropriate to assume a lower rank of the regression matrix.
This will suggest using the approximating property of SVD with a reduced singular value set. Leaving out the components with small singular values may be equivalent to removing the data noise. Taking a matrix $ S_{\text{mod}} $ with $ \hat{r} $ largest singular values while zeroing the remaining ones (see, e.g.,~\cite{trefethen_numerical_1997}) results in a matrix according to (\ref{Eq.SVDmin}):
\begin{equation}
    B_{\text{mod}} = U_{\text{mod}} S_{\text{mod}} V_{\text{mod}}^T
\end{equation}

that has the least matrix $ L_2 $ norm

\begin{equation}
    \norm[\big]{B - B_{\text{mod}}}_2
\end{equation}

out of all existing matrices $ B_{\text{mod}} $ with rank $ \hat{r} $.
The $ L_2 $ matrix norm is induced by the $ L_2 $ vector norm, as defined in Equation~\ref{Eq.L2def}.
The definition (\ref{Eq.L2def}) of the $ L_2 $ matrix norm has an implication for the accuracy of the forecasts with help of $ R $ and $ R_{\text{mod}} $:
\begin{equation}
    \norm[\big]{B - B_{\text{mod}}}_2 = \max_x{\frac{\norm[\big]{\left( B - B_{\text{mod}} \right) x}_2}{\norm[\big]{x}_2}} = \max_x{\frac{\norm[\big]{ y - y_{\text{mod}}}_2}{\norm[\big]{x}_2}}
\end{equation}

The vector norm of the forecast error equal to the square root of the mean square error is obviously minimal for a given norm of the input vector. In other words, the modified, reduced-rank regression matrix has the least maximum forecast deviation from the original regression matrix relative to the norm of the input vector $ x $.

\subsection{Under-determined Problems}\label{sec:undetermined_problems}
A different situation is if the linear regression is under-determined. This is frequently the case in high-dimensional applications such as computer vision and corpus-based semantics---the number of training examples may be substantially lower than the dimensions of the input. The training examples span a subspace of the input vector space. Using this training information, new patterns can only be projected onto this subspace. The projection operator, using the same definition of the input matrix $ X $ as above, is given as:
\begin{equation}
    \hat{x} = X X^{+} x = X \left( X^T X \right)^{-1} X^T x
\end{equation}

This can be viewed as a pattern-specific weighting of training examples by a weight vector $ w_x $

\begin{equation}
    \hat{x} = X w_x^T
\end{equation}

To recall the corresponding output, the same weight vector can be used:

\begin{equation}
    \hat{y} = Y w_x^T = Y X^{+} x = Y \left(X^T X \right)^{-1} X^T x
\end{equation}

This is equivalent to solving the regression problem

\begin{equation}\label{Eq.Runderdet}
    Y = R X
\end{equation}
with help of the pseudo-inverse (see, e.g.,~\cite{kohonen_self-organization_1989}) of X, which is (\ref{Eq.RegrUD}) in the under-determined case.

The regression matrix $ R $ is, as usual, of size $  m \times n $. If the input dimension $ m $ exceeds the number of training examples the regression matrix $ R $ solving Equation~\ref{Eq.Runderdet} is not full rank. Its SVD will exhibit some zero singular values and can be reduced, without a loss of information, to a reduced form:
\begin{equation}
    B_{\text{red}} = U_{\text{red}} S_{\text{red}} V_{\text{red}}^T
\end{equation}

\section{SVD and Linear Networks}\label{sec:svd_and_lnn}
Before establishing the relationship between SVD and nonlinear neural networks, let us consider hypothetical multi-layer networks with linear units of the form $ g \left( x \right) = x$ in the hidden layer.

Suppose a network with one hidden layer of predefined size $ p $ is used to represent a mapping from input $ x $ to output $ y $. Suppose now that the best linear mapping from input $ x $ to output $ y $ is

\begin{equation}
    y = B x
\end{equation}

The best approximation with a rank limitation to $ \hat{r} $ and is, according to (\ref{Eq.SVDmin}):

\begin{equation}
y = U_1 S_1 V_1^T x
\end{equation}

This expression can be viewed as a network with one linear hidden layer of width $ p = \hat{r} $. The weight matrix between the input and the hidden layers is 
\begin{equation}\label{Eq.LinHid}
    V_1^T
\end{equation}

and that between the hidden and the output layers is 

\begin{equation}\label{Eq.LinOut}
    U_1 S_1
\end{equation}

This network has the property of being the best approximation of the mapping from the input to the output between all networks of this size with orthonormal (in the hidden layer) and orthogonal (in the output layer) weight vectors.

This optimality is not strictly guaranteed to be reached if relaxing the orthogonality constraints. The difference between the orthogonal and the non-orthogonal solutions depends on the ratio between the input and the output widths, and on the relative width of the hidden layer in the following way.

How serious this optimality gap may be can be assessed observing the fraction of the number of orthogonality constraints to the number of parameters. If this fraction is small, the number of independent parameters is close to the number of all parameters and the influence of the orthogonality constraints is small.

With hidden layer size $ r $ (equal to the rank of the linear mapping), the total number of constraints is $ r^2 + r $. With $ m < n $ and $ n = q m, q \geq 1 $, the total number of parameters is $ \left( m + q m + 1 \right) r $.
The fraction, and its approximation for realistic values of $ r \gg 1 $ is then

\begin{equation}
    \frac{r^2 + r}{ \left( m + q m + 1 \right) r } = \frac{r + 1}{m + q m + 1} \approx \frac{r}{\left( 1 + q \right) m}
\end{equation}

This fraction decreases with the ratio $ \frac{m}{r} $ (the degree of \emph{feature compression} by the network) and the ratio $ q $.
Since both ratios will usually be large in practical problems of the mentioned domain, the distance to the optimality after relaxing the orthogonality constraints can be expected to be small.

\section{SVD and Initializing Nonlinear Neural Networks}
\begin{figure}
\caption{Plot of Equation~\ref{Eq.Sig2}}\label{Fig.Sig2}
\centering
\includegraphics[height=.45\textheight]{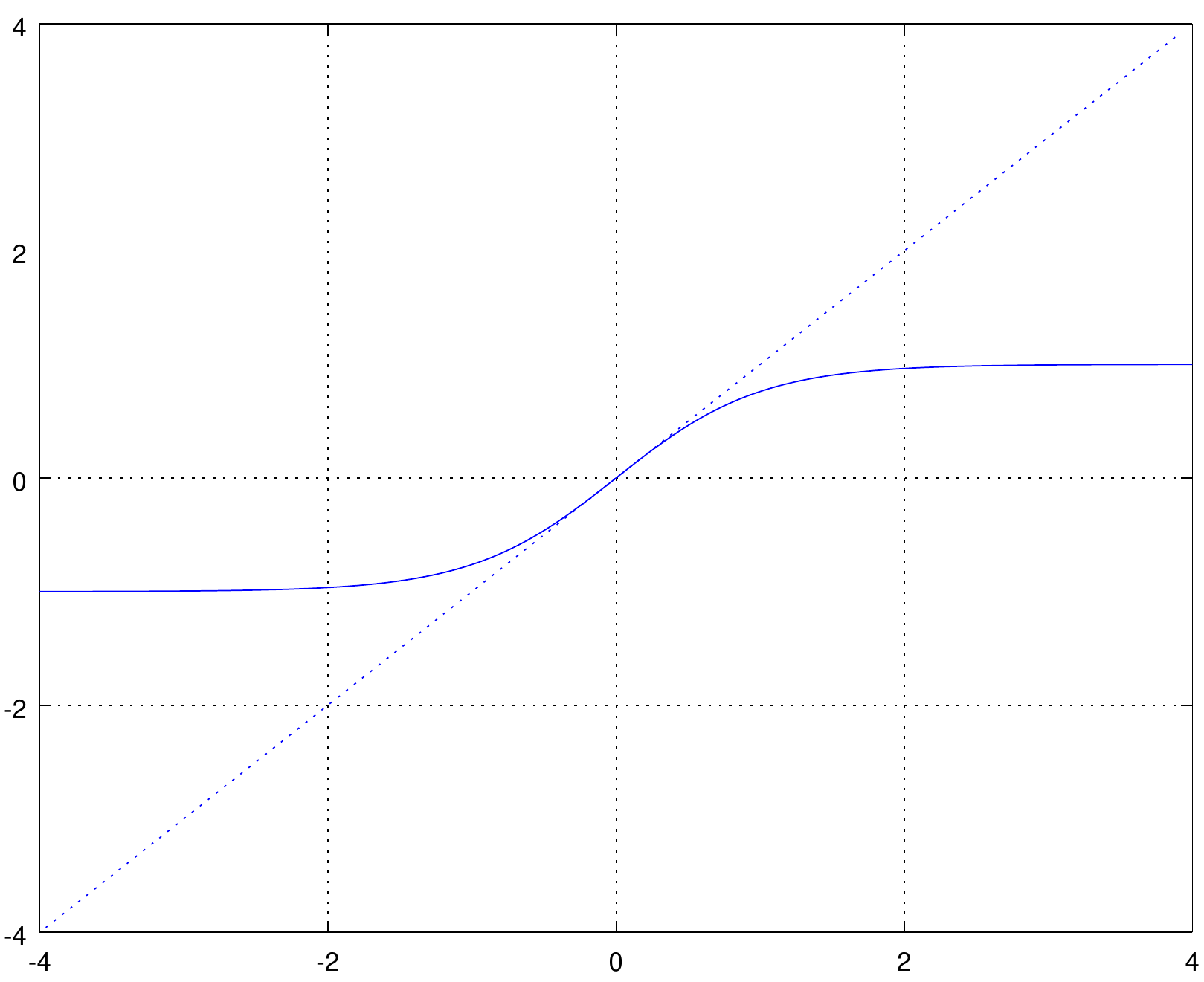}
\end{figure}

Most popular hidden units possess a linear or nearly linear segment. A \emph{sigmoid} unit
\begin{equation}
    s \left( x \right) = \frac{1}{1 + e^{-x}}
\end{equation}

is nearly linear around the point $ x = 0 $ where its derivative is equal to $ 0.25 $. Rescaling this unit to the symmetric form

\begin{equation}\label{Eq.Sig2}
    f \left( x \right) = 2 s \left( 2 x \right) - 1 = \frac{2}{1+e^{-2 x}} - 1
\end{equation}

we obtain a nonlinear function the derivative of which is unity around $ x = 0 $, plotted in Figure~\ref{Fig.Sig2}.

\begin{figure}
\caption{Plot of relative deviation between a linear function and sigmoid}\label{Fig.Sig2vsLin}
\centering
\includegraphics[height=.45\textheight]{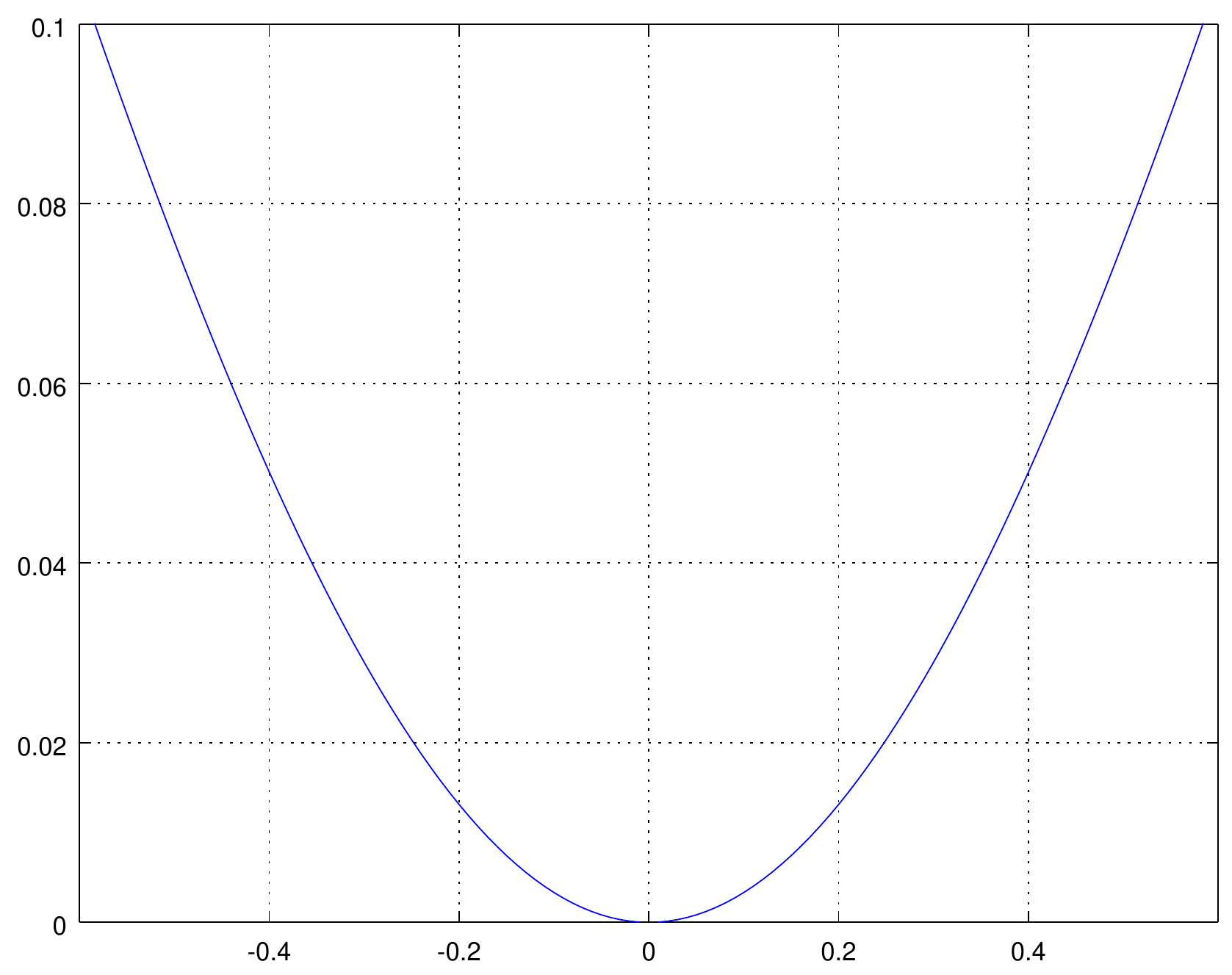}
\end{figure}

Its relative deviation from the linear function $ g \left( x \right) = x $ is below ten percent for $ q \abs{x} < 0.58 $ (see Fig.~\ref{Fig.Sig2vsLin}). So, a neural network with one hidden layer using the sigmoid activation function (\ref{Eq.Sig2}) behaves like a linear network for small activation values of the hidden layer.

This fact can be used for finding a good initial guess of parameters of a nonlinear neural network with a single hidden layer. The in-going weights into the hidden and output layers are (\ref{Eq.LinHid}) and (\ref{Eq.LinOut}), respectively.

\section{Computing Experiments}\label{sec:experiments}
A series of computing experiments have been carried out to assess the real efficiency of using SVD as a generator of the initial state of neural network parameters. To provide a meaningful interpretation of the mean square figures attained, all problems have been deliberately defined to have a minimum at zero. To justify the use of the hidden layer as a feature extractor, its width should be smaller than the minimum of the input and output sizes. The dimensions have been chosen so that the full regression would be under-determined (as typical for the application types mentioned above), but the use of a hidden layer with a smaller width makes it slightly over-determined. So, the effect of overfitting, harmful for generalization, is excluded.

The software used for SVD computation was the Python module \emph{SciPy}~\cite{jones_scipy:_2001}. Neural networks were optimized by several methods implemented in the popular framework \emph{Keras}~\cite{chollet_keras_2015}: \emph{Stochastic Gradient Descent} (SGD), selected because of its widespread use, as well as \emph{Adadelta} and \emph{RMSprop}, which seem to be the most efficient ones for the problems considered.

Typical Keras-methods are first order and there is a widespread opinion in the neural network community that second-order methods are not superior to the first order ones. However, there are strong theoretical and empirical arguments in favor of the second-order methods from numerical mathematics. So the conjugate gradient method (CG), as implemented in SciPy, has also been applied. Since the SciPy/Keras interface failed to work for this method\footnote{Since we use TensorFlow as Keras' backend execution engine, the resulting computation graph would have been cut into two different executions for each optimization step which causes a too high computational overhead.}, efficient Keras-based network evaluation procedures could not be used. So, for the largest problems, the CG method had to be omitted. 

The performance of the optimization methods has been compared with the help of the number of gradient calls.
All methods have been used with the default settings of Keras and SciPy.

Three problem sizes denoted as $ A $, $ B $, and $ C $ have been used. Using different size classes will make it possible to discern possible dependencies on the problem size if there are any. The largest size of class $ C $ is still substantially below that of huge networks such as \emph{VGG-19}~\cite{simonyan_very_2014} used in image classification. The computing effort for making method comparison with such huge sizes would be excessive for the goals of this study. However, we believe the size is sufficient for showing trends relevant for very large network sizes.

The three size classes are characterized by their input and output dimensions as well as by the size of the training set. The concrete network sizes, parameter numbers, and numbers of constraints (output values to be reached times the number of training examples) are given in Table~\ref{Tab.ProblemDef}. The column ``\#\ constraints'' shows the number of constraints imposed by the reference outputs to be fitted. It is the product of the output dimension and the training set size. Comparing the number of constraints with the number of parameters defines the over-determination or the under-determination of the problem (e.g., a problem with more constraints than parameters is over-determined).

The results for the different size classes are given in Table~\ref{Tab.SGDres}. 
For each network architecture, three different parametrizations with corresponding training sets have been generated, all with a known mean square error minimum of zero. For every variant, an SVD has been computed and used to determine the network initialization. For comparison, five different random network initializations have been generated. The results below are geometric means of minima reached (means from three optimization runs for SVD initializations, and means from $ 3 \cdot 5 = 15$ runs for random initializations).

\begin{table}
\caption{Test problem definitions}\label{Tab.ProblemDef}
\centering
\begin{tabular*}{\textwidth}{c @{\extracolsep{\fill}} rrrr rr}
	\toprule
	Type & \#input    &  \#output &  \#hidden & \#training & \#parameters & \#constraints \\ \midrule
	 A   & \num{ 100} & \num{ 50} & \num{ 20} & \num{  80} & \num{  3070} &  \num{  4000} \\
	 B   & \num{ 300} & \num{150} & \num{ 60} &  \num{240} & \num{ 27210} &  \num{ 36000} \\
	 C   & \num{1000} & \num{500} & \num{200} &  \num{800} & \num{300700} &  \num{400000} \\ \bottomrule
\end{tabular*}
\end{table}

\begin{table}
\caption{Mean square minima reached by various optimization methods with random and SVD-based initial parameter sets}\label{Tab.SGDres}
\centering
\begin{tabular*}{\textwidth}{@{\extracolsep{\fill}} ll rr rr rr}
	\toprule
	Algorithm & Init.  &                \multicolumn{2}{c}{size class $ A $}                &                \multicolumn{2}{c}{size class $ B $}                &                \multicolumn{2}{c}{size class $ C $}                \\
	          &        & \scriptsize{\#iter.} & \scriptsize{$ F_{opt} \times 10^{-3} $} & \scriptsize{\#iter.} & \scriptsize{$ F_{opt} \times 10^{-3} $} & \scriptsize{\#iter.} & \scriptsize{$ F_{opt} \times 10^{-3} $} \\ \midrule
	SVD       & ---      & ---                    &                            \num{10.200} & ---                    &                            \num{10.559} & ---                    &                           \num{10.814} \\
	SGD       & Random & \num{2000}           &                            \num{30.361} & \num{2000}           &                            \num{90.402} & \num{2000}           &                           \num{177.095} \\
	RMSprop   & Random & \num{2000}           &                            \num{0.040} & \num{2000}           &                            \num{0.096} & \num{2000}           &                           \num{0.260} \\
	Adadelta  & Random & \num{2000}           &                            \num{1.290} & \num{2000}           &                            \num{6.748} & \num{2000}           &                           \num{31.076} \\
	CG        & Random & \num{637}           &                            \num{0.002} & \num{ 821}           &                            \num{0.012} & ---                    &                                       --- \\
	SGD       & SVD    & \num{2000}           &                            \num{1.779} & \num{2000}           &                            \num{4.145} & \num{2000}           &                           \num{7.254} \\
	RMSprop   & SVD    & \num{2000}           &                            \num{0.030} & \num{2000}           &                            \num{0.085} & \num{2000}           &                           \num{0.248} \\
	Adadelta  & SVD    & \num{2000}           &                            \num{0.062} & \num{2000}           &                            \num{0.511} & \num{2000}           &                           \num{2.086} \\
	CG        & SVD    & \num{316}           &                            \num{0.000} & \num{233}           &                            \num{0.021} & ---                    &                                       --- \\ \bottomrule
\end{tabular*}
\end{table}

The results of four optimization methods are given in the randomly initialized variant and in the variant initialized with help of the SVD solution. 

The first row, labeled with algorithm ``SVD'', shows the minima reached by the SVD solution without any subsequent optimization. It is obvious that the SVD-based initialization is pretty good. Its mean square error minimum is substantially better than the weakest Keras-method \emph{SGD} with random initialization. For the largest problem size class, SVD without optimization is also superior to Adadelta with random initialization.

An SVD-based initialization with a subsequent optimization lets \emph{SGD} reach an acceptable minimum, with even better results using \emph{Adadelta}.
The best Keras-method, \emph{RMSprop}, was clearly inferior to the conjugate gradient (\emph{CG}), although CG stopped the optimization substantially earlier that the fixed iteration number of RMSprop. For both these methods, the improvement by SVD-based initialization was weak (for CG only in the number of iterations). This is not unexpected: good optimization methods are able to find the representations similar to the SVD by themselves, solving a closely related problem with a different numerical procedure.

\section{Conclusion and Discussion}\label{sec:conclusion}
SVD constitutes a bridge between the linear algebra concepts and multi-layer neural networks---it is their linear analogy. Besides this insight, it can be used as a good initial guess for the network parameters. The quality of this initial guess may be, for large problems, better than weakly performing (but widely used) methods such as SGD ever reach.

It has to be pointed out that as long as the network uses nonlinear hidden units, simply using this initial guess as ultimate network parameters makes little sense: it would be preferable to make the units linear, and use the SVD matrices directly to represent the desired input-output mapping.

Unfortunately, there seems to be no analogous generalization for networks with multiple hidden layers.
With a hidden layer sequence of monotonically decreasing width (for example, from the input towards the output) it would be possible to proceed iteratively, by successively adding hidden layers of decreasing width. 

The procedure would start by defining the first hidden layer $ z_1 $ (the one with the largest dimension) and initializing its weights with the help of SVD\@. Then, the following iterations over the desired number of hidden layers would be performed:
\begin{enumerate}
    \item Analyzing the mapping between the output of the last hidden layer considered and the output layer $ z_i \rightarrow y $ (with $ z_0 = x $) using SVD\@.
    \item Finding an initial guess of parametrization for the incoming weights to $ z_{i + 1} $.
    \item Optimizing the weights of such extended nonlinear network by some appropriate optimization method.
\end{enumerate}
This is a formal generalization of the procedure for the network with a single hidden layer $ z_1 $, as presented above.

However, it is difficult to find a founded justification for this procedure, as it is equally difficult to find a founded justification for using multiple fully connected hidden layers at all---although there seems to be empirical evidence in favor of this. Of course, there are good justifications for using special architectures such as convolutional networks, which are motivated, e.g., by spatial operators in image processing.

\bibliographystyle{splncs04}
\bibliography{ms}

\end{document}